\documentclass[11pt]{article}

\usepackage[T1]{fontenc}
\usepackage[utf8]{inputenc}
\usepackage{lmodern}
\usepackage[margin=1in]{geometry}
\usepackage{microtype}
\usepackage{amsmath,amssymb,amsthm}
\usepackage{mathtools}
\usepackage{booktabs}
\usepackage{array}
\usepackage{graphicx}
\usepackage{hyperref}
\usepackage{xcolor}

\hypersetup{
  colorlinks=true,
  linkcolor=blue!50!black,
  citecolor=blue!50!black,
  urlcolor=blue!50!black,
  pdftitle={Aligning Cellular Sheaves with Classifier Attention for
            Interpretable Weakly-Supervised Pathology Localization},
  pdfauthor={Devansh Lalwani, Swapnil Bhat, Maulik Shah}
}

\newcommand{\R}{\mathbb{R}}
\newcommand{\Fcal}{\mathcal{F}}
\newcommand{\norm}[1]{\left\lVert#1\right\rVert}

\title{\bfseries Aligning Cellular Sheaves with Classifier Attention\\
       for Interpretable Weakly-Supervised\\
       Pathology Localization}

\author{%
  Devansh Lalwani\thanks{Corresponding author: \texttt{founders@turocrates.ai}} \quad
  Swapnil Bhat \quad
  Maulik Shah \\[4pt]
  \normalsize\itshape Turocrates AI Private Limited, Mumbai, India
}

\date{}

\begin{document}
\maketitle

\begin{abstract}
Weakly-supervised classification of whole-slide images with
attention-based multiple instance learning (ABMIL) on top of
foundation features now reaches near-saturation on Camelyon16
slide-level performance, but the corresponding attention maps are an
imperfect localization signal: in clinical interpretation, a model
that classifies correctly without firing on the actual lesion is hard
to trust. We address this gap with cellular sheaves, which equip each
vertex and edge of a graph with a finite-dimensional vector space and
consistent linear maps between them, providing a principled way to
detect local disagreement on graph-structured data. We apply cellular
sheaves to weakly-supervised tumour localization on whole-slide
images, combining a sheaf disagreement field with ABMIL. The natural
training objective, encouraging consistency between similar features,
produces a disagreement field that tracks tissue-level texture rather
than diagnostic content. We propose \emph{attention-conditional
consistency}, which uses the classifier's own attention to define
which neighbouring patches should agree. Joint training of the
classifier and the sheaf under this objective produces a disagreement
field with patch-level AUC $0.940$ on Camelyon16 and raises the
attention head from its ABMIL-alone level of $0.717$ to $0.953$. A
two-stage ablation with the classifier frozen at its ABMIL values
reaches only $0.727$ on the disagreement field and leaves attention
at $0.717$, confirming that the gain comes from the projector
co-adapting under both objectives, not from the loss change in
isolation. The trained model transfers without retraining to
annotated slides from Camelyon17, maintaining $\Delta$ AUC
$0.932 \pm 0.083$ and attention AUC $0.955 \pm 0.099$. The result is
an attention map and a sheaf-disagreement map that fire on the same
diagnostic regions, giving the clinician two complementary
explanations for each slide-level prediction.
\end{abstract}

\section{Introduction}\label{sec:intro}

Whole-slide images (WSIs) of histology sections are large
multi-resolution scans, often exceeding 100{,}000 pixels per side.
Slide-level diagnostic labels (``this lymph node contains metastatic
carcinoma'') are routinely available; pixel-level annotations are
not. Multiple-instance learning (MIL) treats each slide as a bag of
patches and learns to predict the slide label from patch-level
features, with attention pooling indicating which patches drove the
prediction. Attention-based MIL (ABMIL) on top of foundation-model
features such as UNI now reaches near-saturation on Camelyon16
slide-level classification.

Reaching saturation on slide-level AUC, however, does not solve the
clinical problem. A pathologist reading the model's output needs to
see \emph{why} a slide was called positive: which region triggered
the prediction, and how confident the model is locally. ABMIL
attention provides a useful proxy but is a known weak localizer. The
attention scalar for each patch is a function of that patch alone
(through a small gated network), with no explicit modelling of how
patches relate to their spatial neighbours. Tumour regions in lymph
node biopsies are spatially coherent: a tumour-bearing patch is
usually surrounded by other tumour-bearing patches, and the boundary
between tumour and normal tissue is the most diagnostically
informative region of the slide. A localization signal that
explicitly models neighbourhood consistency should produce a more
interpretable map for clinical review.

Cellular sheaves provide such a signal. A cellular sheaf on a graph
attaches a vector space to each vertex and each edge, with linear
restriction maps from incident vertices to edges; the sheaf
Laplacian then yields a per-vertex disagreement quantity that
measures how much each vertex deviates from local consistency. The
construction is symmetric, gauge-invariant, and parameterisable, so
it can be learned end-to-end alongside the classifier. Applied to a
patch graph, the sheaf disagreement field becomes a candidate
localization signal that is topological by construction.

The natural recipe, train the sheaf to make feature-similar patches
agree, produces a disagreement field that tracks general tissue
texture rather than diagnostic content. Our contribution is the
observation that the consistency target should be the classifier's
own attention, not raw feature similarity. We introduce
\emph{attention-conditional consistency}, a training objective on
sheaf restrictions that uses per-edge attention differences as
supervision: edges between patches the classifier weights similarly
should agree, edges across attention discontinuities should
disagree. Joint training of the classifier and the sheaf under this
objective produces a sheaf disagreement field whose AUC approaches
that of the attention head as a patch-level localizer, and, more
surprisingly, sharpens the attention map itself.

The combined output, two heads producing localization maps that fire
on the same regions but for different mathematical reasons, is the
interpretability contribution: a clinician can ask both ``where did
the classifier focus?'' and ``where does the tissue itself signal a
transition?'' and get answers that should agree on a correctly
predicted tumour, and disagree informatively when the prediction is
uncertain.

\section{Background}\label{sec:background}

\subsection{Cellular sheaves on graphs}

Let $G = (V, E)$ be an undirected graph. A cellular sheaf $\Fcal$ on
$G$ assigns to each vertex $v \in V$ a vector space $\Fcal(v) =
\R^{d}$, to each edge $e \in E$ an edge stalk $\Fcal(e) = \R^{d'}$,
and to each incidence $v \trianglelefteq e$ a linear restriction map
$\Fcal_{v \trianglelefteq e} : \Fcal(v) \to \Fcal(e)$.

Each edge $e = (u,v)$ is given an arbitrary but fixed orientation.
The coboundary map $\delta$ collects per-edge discrepancies:
\[
\delta_e \;=\; \Fcal_{u \trianglelefteq e}\, x_u \;-\;
              \Fcal_{v \trianglelefteq e}\, x_v .
\]
The sheaf Laplacian is defined as $L_{\Fcal} = \delta^{\!\top}\delta$,
which is symmetric and positive semidefinite by construction. Its
action on the $v$-th block is
\[
(L_{\Fcal}\, x)_v \;=\;
  \sum_{e:\, v = u(e)}\!\Fcal_{v \trianglelefteq e}^{\!\top}\,\delta_e
  \;-\!\!\sum_{e:\, v = v(e)}\!\Fcal_{v \trianglelefteq e}^{\!\top}\,\delta_e ,
\]
where the sign reflects the orientation convention. The kernel of
$L_{\Fcal}$ is exactly the space of global sections, i.e., the
assignments $x$ for which all restriction maps agree. We use the
per-vertex norm of the Laplacian image,
\[
\Delta_v \;=\; \norm{(L_{\Fcal}\, x)_v}_2 ,
\]
as a per-vertex disagreement signal.

\subsection{Attention-based MIL with foundation features}

ABMIL embeds each patch through a small projection, computes a gated
attention scalar for each patch, pools features by attention-weighted
sum, and classifies the slide. With UNI patch features (a
1024-dimensional ViT-L/16 representation pretrained on a large
pathology corpus) ABMIL attention is already a strong slide-level
classifier on Camelyon16 (${\sim}0.95$ AUC). Treating the attention
scalars as a localization map gives a competitive but imperfect
patch-level signal.

\section{Method}\label{sec:method}

\subsection{Patch graph and low-rank sheaf}

Each slide is processed as a graph $G = (V, E)$ where vertices are
tissue patches at 128\,\textmu m physical resolution. Edges connect
each patch to its 8 spatial nearest neighbours by patch centroid.
Each vertex stalk $\Fcal(v) = \R^{d}$ carries the projected UNI
feature with $d = 256$, and each edge stalk has dimension $d' = 64$.

Restriction maps are produced by a hypernetwork conditioned on the
two endpoint features and their spatial offset, in low-rank form
$\Fcal_{v \trianglelefteq e} = A B^{\!\top}$ with $A \in \R^{d'
\times r}$, $B \in \R^{d \times r}$, and rank $r = 8$. The low-rank
parameterisation reduces parameter count and forces each restriction
map to identify a small set of directions in feature space along
which the two endpoint patches are compared.

\subsection{Three properties of the construction}

\begin{itemize}
\item \textbf{Positive semidefiniteness.} The sheaf Laplacian
  $L_{\Fcal} = \delta^{\!\top}\delta$ is PSD by construction, so the
  disagreement $\Delta_v \ge 0$ with equality iff the section is
  consistent at vertex $v$.

\item \textbf{Gauge invariance.} Replacing each restriction map by
  an orthogonal change of basis on the edge stalks leaves $\Delta_v$
  unchanged. The disagreement is a property of the sheaf data, not
  of the parameterisation.

\item \textbf{Geometric meaning of low rank.} With rank $r$, the
  local disagreement at a vertex sees only the projection of the
  feature onto an $r$-dimensional subspace, selected by the
  hypernetwork per edge. Larger $r$ gives more capacity per edge; we
  use $r = 8$, which is enough for this task.
\end{itemize}

\subsection{Joint training: classifier and sheaf}

The model has a single shared input projector. From the projected
features, the classifier head computes gated attention and a
slide-level logit; the sheaf head computes the disagreement field
$\Delta$. Both heads are exposed simultaneously through one forward
pass. Training jointly optimises classification cross-entropy on the
slide label, a consistency loss on the sheaf, and a weak
orthogonality regulariser on the low-rank factors.

\subsection{Attention-conditional consistency}\label{sec:acc}

The choice of consistency loss determines what the disagreement
field learns. We propose the following. Let $a$ be the (detached)
attention vector produced by the classifier head. On each edge
$(u,v)$ we form $|a_u - a_v|$: the gap between the classifier's
importance judgements at the two endpoints. We split edges by
per-slide percentile thresholds: low-gap edges (bottom 30\%) are
treated as agreeing, high-gap edges (top 30\%) as disagreeing.

On agreeing edges we minimise $\norm{\delta_e}^2$: the sheaf is
encouraged to give consistent restrictions where the classifier
already agrees. On disagreeing edges we encourage $\norm{\delta_e}^2$
above a margin: the sheaf is encouraged to express disagreement
exactly where the classifier itself sees a boundary. The total loss
is the weighted sum of these two terms.

Attention is detached when computing this loss, so no gradient flows
from the consistency loss back into the classifier through the sheaf
branch. Updates to the projector still come from both objectives,
allowing the projector to find a representation that serves both
heads. The classifier head is not directly modified by the sheaf
objective; the sheaf adapts to whatever attention pattern the
classifier produces.

The intuition: a useful per-patch disagreement signal should fire
exactly where the classifier's per-patch judgements transition.
Attention-conditional consistency makes this alignment explicit at
training time, rather than hoping it emerges from
feature-similarity-based consistency.

\section{Experiments}\label{sec:experiments}

\begin{figure*}[t]
\centering
\includegraphics[width=0.95\textwidth]{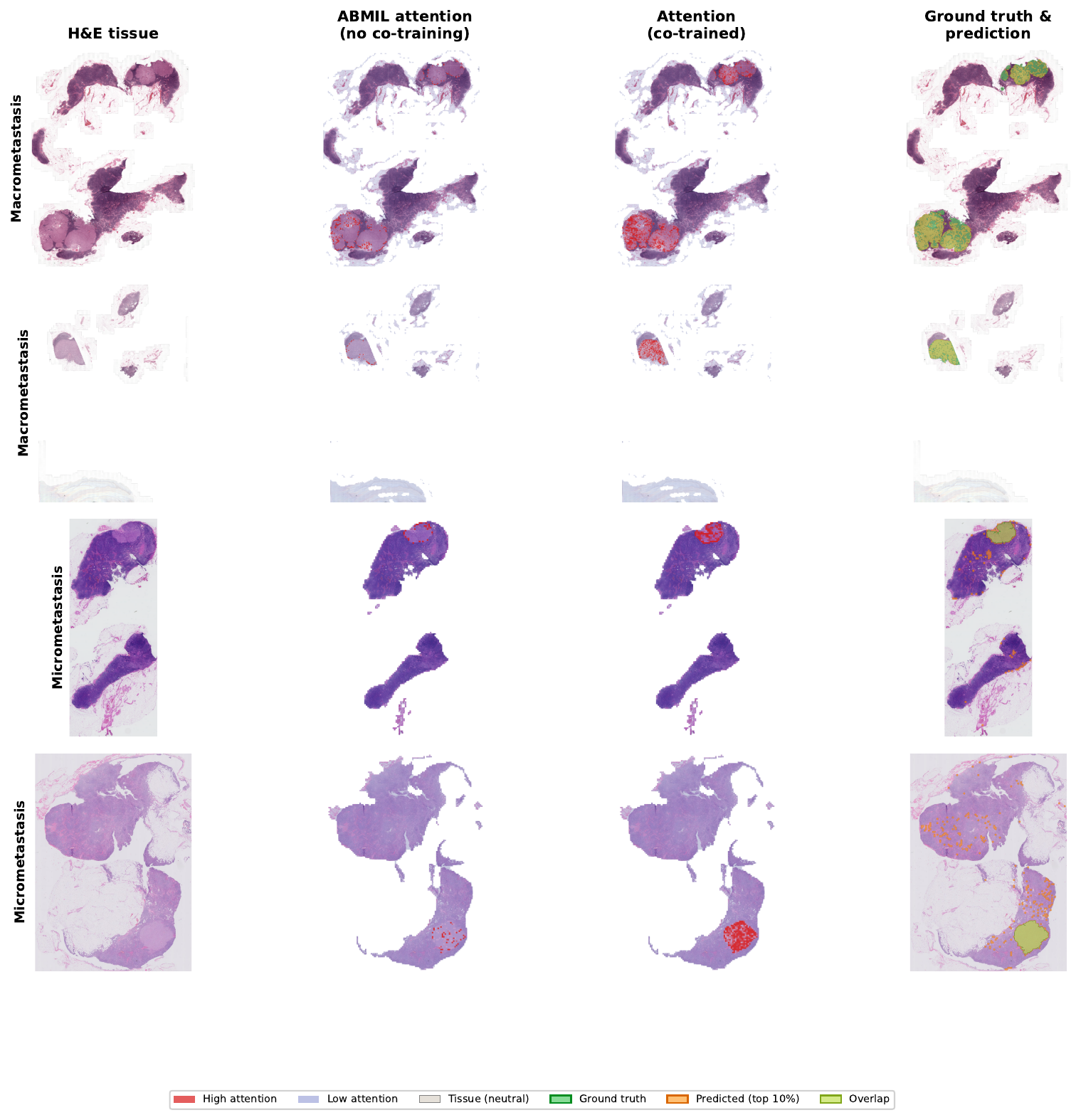}
\caption{Attention heatmaps on four Camelyon17 slides (two
macrometastases, two micrometastases), evaluated without any
retraining on C17. Each row shows a different slide. \textbf{Column
1:}~H\&E tissue. \textbf{Column 2:}~ABMIL attention without sheaf
co-training (frozen-classifier model): attention fails to localize
on tumour regions. \textbf{Column 3:}~Attention from the jointly
trained Sheaf-MIL-Loc model: attention concentrates on
tumour-bearing regions. Red indicates high attention; blue indicates
low attention. \textbf{Column 4:}~Ground-truth tumour mask (green)
overlaid with the predicted tumour region (orange, top-10\%
co-trained attention patches); yellow indicates overlap between
ground truth and prediction. The co-trained model localizes both
large macrometastases and small micrometastatic deposits, while
ABMIL-only attention does not.}
\label{fig:heatmaps}
\end{figure*}

\subsection{Datasets and setup}

We use Camelyon16 for training and primary evaluation, and
Camelyon17 for external validation. Camelyon16 contains 270 training
slides (110 tumour, 160 normal) and 129 test slides (49 tumour, 80
normal). Camelyon17 contains 50 mask-annotated slides drawn from 5
medical centres in the Netherlands.

Patch features are extracted with UNI~\cite{chen2024uni}
(ViT-L/16, 1024-dim) at 128\,\textmu m physical resolution using the
same tissue-detection, tiling, and inference pipeline for both
datasets. $k$NN graphs use $k = 8$ in spatial coordinates. Training
uses AdamW with learning rate $10^{-4}$, weight decay $10^{-5}$,
batch size 1 slide, gradient clipping at norm $1.0$, and 40 epochs
with the first 5 as classifier-only warmup.

\textbf{Limitations and future work.}
We compare against ABMIL~\cite{ilse2018abmil} as the sole MIL
baseline. Patch-level localization comparisons with
CLAM~\cite{lu2021clam}, DSMIL~\cite{li2021dsmil}, and TransMIL
under the same UNI feature protocol, multi-seed variance estimates,
and a sensitivity analysis over architectural hyperparameters
($r{=}8$, $k{=}8$, $d'{=}64$) are planned as future work.

\subsection{Camelyon16 localization}\label{sec:c16}

Table~\ref{tab:c16} reports per-slide patch-level AUC averaged
across the 49 tumour test slides, for the trained Sheaf-MIL-Loc
model with attention-conditional consistency, the two-stage
frozen-classifier ablation, and ABMIL attention as the baseline
localizer. Numbers are mean $\pm$ SD across slides; $\Delta$ is the
sheaf disagreement field; attention is the gated-attention head.

\begin{table}[h]
\centering
\small
\setlength{\tabcolsep}{4pt}
\begin{tabular}{@{}lcccc@{}}
\toprule
\textbf{Method} & $\boldsymbol{\Delta}$ \textbf{AUC}
                & \textbf{Attn.\ AUC}
                & $\boldsymbol{\Delta}$ \textbf{p@50}
                & \textbf{Attn.\ p@50} \\
\midrule
ABMIL (no sheaf)          & ---   & 0.717 & ---   & 0.569 \\
Sheaf-MIL-Loc (joint)     & 0.940 & 0.953 & 0.521 & 0.632 \\
Two-stage (frozen cls.)   & 0.727 & 0.717 & 0.329 & 0.569 \\
\bottomrule
\end{tabular}
\caption{Camelyon16 patch-level localization on 49 tumour test
slides (single seed). Joint training raises the attention AUC from
$0.717$ to $0.953$. The two-stage variant with a frozen classifier
leaves attention at $0.717$ and gives only a modest $\Delta$. The
gain comes from the projector co-adapting under both objectives.
Note that $\Delta$ prec@50 ($0.521$) is lower than attention
prec@50 ($0.632$): the disagreement field's strength is in
whole-slide ranking (AUC) rather than top-$k$ precision, likely
because $\Delta$ fires on tumour--normal boundaries rather than
tumour centres.}
\label{tab:c16}
\end{table}

The attention-conditional consistency objective produces a
disagreement field whose AUC approaches that of the attention head
($0.940$ vs $0.953$); Figure~\ref{fig:heatmaps} illustrates this
on representative Camelyon17 slides. The $\Delta$ field is stronger
in whole-slide ranking (AUC) than in top-$k$ precision: $\Delta$
prec@50 is $0.521$ versus attention prec@50 of $0.632$, reflecting
the fact that $\Delta$ activates on tumour--normal transition zones
rather than tumour centres.

The more striking effect is on the attention head itself: pure
ABMIL with no sheaf branch reaches patch-level AUC $0.717$,
consistent with the published consensus that ABMIL attention is a
weak localizer; the same ABMIL architecture trained jointly with
the sheaf and the attention-conditional consistency loss reaches
$0.953$, an improvement of $0.236$ on the same test slides. The
two-stage variant isolates the cause. With the classifier frozen at
its ABMIL-trained values, training only the sheaf hypernet on top
of the new loss raises $\Delta$ to $0.727$ but leaves attention at
$0.717$ by construction. The $0.236$ attention gain in the joint
setting therefore comes from the projector receiving gradients from
both the classification head and the sheaf consistency objective,
not from the loss change in isolation. The sheaf branch is acting
as a structured regulariser on the projector: the representation
that minimises attention-conditional consistency also produces
sharper attention for slide-level classification.

We also tried a naive feature-similarity consistency loss, where
edges between feature-similar patches are pulled toward zero
disagreement. This produces a degenerate $\Delta$ field (AUC
$0.575$) that tracks tissue-level texture rather than diagnostic
content, while the attention head converges to roughly the same
value as ABMIL alone ($0.717$). The choice of consistency target is
therefore as important as the decision to train the sheaf jointly
with the classifier.

\subsection{Camelyon17 evaluation}\label{sec:c17}

We evaluate the trained Camelyon16 model on Camelyon17 without any
retraining or fine-tuning. Camelyon17 provides 50 slides with
publicly available pixel-level annotations from the official
challenge release. These slides were collected at medical centres
distinct from those producing the Camelyon16 dataset and with
different scanner generations. On these 44 evaluable tumour slides,
the disagreement field reaches $\Delta$ AUC $0.932 \pm 0.083$ and
attention AUC $0.955 \pm 0.099$ (mean $\pm$ SD across slides),
closely matching the in-domain Camelyon16 results of $0.940$ and
$0.953$ (see Figure~\ref{fig:heatmaps} for representative
examples). That the C17 attention AUC ($0.955$) is marginally higher
than the C16 value ($0.953$) is within the per-slide noise (C17 SD
$= 0.099$); we do not interpret this as a genuine gain from domain
shift. The trained model transfers to Camelyon17 without measurable
degradation, despite the change in acquisition protocol.

\subsection{Diagnostic interpretation}

We measure per-slide Spearman correlations between $\Delta$ and four
reference quantities: the patch feature norm $\norm{x}$, the mean
cosine similarity of each patch to its $k$NN neighbours, the
$k$NN-graph hop-distance to the nearest opposite-label patch, and
the binary tumour label. With attention-conditional consistency,
mean correlations across tumour test slides show that $\Delta$
tracks neighbourhood feature dissimilarity (Spearman $\rho \approx
-0.85$ with mean neighbour cosine similarity, by construction of the
loss) and is positively associated with the tumour label. The
disagreement field is functionally distinct from attention: the two
heads are not redundant readouts of the same underlying signal.

\section{Discussion}\label{sec:discussion}

Cellular sheaves provide a natural language for per-patch
disagreement on graph-structured WSI data, but the natural training
objective, feature-similarity consistency, does not by itself
produce a useful localization signal. The fix is to align the
consistency objective with the classifier's attention rather than
with raw feature similarity. Attention is a learned signal that
already encodes diagnostic relevance; using its edge-wise differences
to define agreement and disagreement gives the sheaf a meaningful
target. The framing is consistent with concurrent work on aligning
classifier prediction with attention saliency in MIL pipelines, but
introduces a different alignment mechanism: a structured loss on a
topological field rather than a parameter-sharing constraint between
heads.

The two-stage ablation rules out a possible alternative explanation.
If the gain came purely from the new loss, training only the sheaf
hypernet on top of a frozen classifier would reproduce the joint
result. It does not: the two-stage $\Delta$ AUC of $0.727$, while
improved over baseline, falls well short of the joint $0.940$. Both
heads need to co-adapt during training; the projector ends up
serving classification and sheaf consistency simultaneously, and
that joint optimum is what produces the result. A practical
consequence is that the sheaf branch can be added to an existing
ABMIL pipeline at small additional cost and improves not only
localization but the attention map itself.

The Camelyon17 evaluation confirms that the result is not specific
to Camelyon16 tissue characteristics. The trained model transfers to
annotated Camelyon17 slides without retraining, achieving $\Delta$
AUC $0.932 \pm 0.083$ and attention AUC $0.955 \pm 0.099$, matching
in-domain numbers despite the different acquisition pipeline. The
difference between C16 and C17 attention AUCs ($0.953$ vs $0.955$)
is well within the per-slide standard deviation and we do not
consider it meaningful.

The interpretability outcome is twofold. First, the attention map
produced by the co-trained model is itself a sharper localization
signal than the attention map produced by ABMIL trained alone,
requiring no architectural change. Second, the sheaf disagreement
field gives a complementary view, derived from the topology of the
patch graph rather than from the classifier directly. On a correctly
predicted positive slide, the two fields fire on the same regions,
providing concordant explanations for the prediction. On a slide
where the two fields disagree, the disagreement is itself a useful
signal: the classifier's attention may have been driven by a
confounder that the topological view does not corroborate, or vice
versa. We leave systematic exploration of this calibration use case
to clinical follow-up work.

The framework is conservative in important ways. We do not modify
the sheaf mathematics, only the consistency loss. The classifier
remains a standard ABMIL head, and feature extraction uses publicly
available UNI weights without any fine-tuning. The architecture thus
inherits the efficiency and reproducibility properties of the
underlying ABMIL+UNI pipeline while adding a localization-aware head
trained at small additional cost.

\section{Conclusion}\label{sec:conclusion}

We presented an application of cellular sheaf neural networks to
weakly-supervised tumour localization on whole-slide images, with
explicit attention to the interpretability requirements of clinical
pathology. The key methodological contribution is
attention-conditional consistency, a training objective that aligns
the sheaf disagreement field with the classifier's attention. With
this objective, joint training of classifier and sheaf produces a
disagreement field that approaches attention as a patch-level
localizer on Camelyon16 ($\Delta$ AUC $0.940$), sharpens the
attention map itself (from $0.717$ to $0.953$ patch-level AUC), and
transfers without retraining to annotated Camelyon17 slides
($\Delta$ AUC $0.932 \pm 0.083$). The two-stage ablation
establishes that joint optimisation is essential. The end-result is
a pair of complementary localization maps that fire on the same
regions for different mathematical reasons, supporting clinical
interpretation of individual predictions.

The same construction extends naturally to other weakly-supervised
localization problems in pathology: detection of mitotic figures,
identification of high-grade regions in graded cancers, and
characterisation of pre-malignant lesions. We leave these to future
work.

\section*{Acknowledgements}

We thank the organizers of the Camelyon16 and Camelyon17 challenges
for making the datasets and ground-truth annotations publicly
available, and the Mahmood Lab for releasing the UNI foundation
model.

\end{document}